# A Framework for Multi-View Classification of Features

Khalil Taheri[a], Hadi Moradi[a,b]*, Mostafa Tavassolipour[a]

[a] Machine Intelligence and Robotics Department, School of Electrical and Computer Engineering, College of Engineering, University of Tehran, Tehran, 11155-4563, Iran, k.taheri@ut.ac.ir, moradih@ut.ac.ir, tavassolipour@ut.ac.ir
[b] Intelligent System Research Institute (ISRI), SKKU, Suwon, 16419, South Korea,
*(Corresponding author: moradih@ut.ac.ir, +98 21 61114960)

**Abstract**

One of the most important problems in the field of pattern recognition is data classification. Due to the increasing development of technologies introduced in the field of data classification, some of the solutions are still open and need more research. One of the challenging problems in this area is the curse of dimensionality of the feature set of the data classification problem. In solving the data classification problems, when the feature set is too large, typical approaches will not be able to solve the problem. In this case, an approach can be used to partition the feature set into multiple feature sub-sets so that the data classification problem is solved for each of the feature subsets and finally using the ensemble classification, the classification is applied to the entire feature set. In the above-mentioned approach, the partitioning of feature set into feature sub-sets is still an interesting area in the literature of this field. In this research, an innovative framework for multi-view ensemble classification, inspired by the problem of object recognition in the multiple views theory of humans, is proposed. In this method, at first, the collaboration values between the features is calculated using a criterion called the features collaboration criterion. Then, the collaboration graph is formed based on the calculated collaboration values. In the next step, using the community detection method, graph communities are found. The communities are considered as the problem views and the different base classifiers are trained for different views using the views' corresponding training data. The multi-view ensemble classifier is then formed by a combination of base classifiers based on the AdaBoost algorithm. The simulation results of the proposed method on the real and synthetic datasets show that the proposed method increases the classification accuracy.

# 1. Introduction

One of the most important problems in the field of pattern recognition is data classification [1]. Data classification has been studied for many years and different methods have been proposed to solve it [2]. These methods can be referred to as different methods for feature extraction, proposing different classifiers, and evaluating their performance [2]. In [3] Zhang considers fundamental development in classification using artificial networks. Specifically, issues such as posterior probability estimation, communication between traditional and neural segments, a compromise between learning and generalization in classification and selection of feature variables in this research have been studied. The purpose of this study is to investigate and stimulate the interest of others for further development and effort. According to Zhang's work, due to the ever-increasing development of data in the field of data classification, some of the solutions are still open and need more research. One of the open problems in this area is the curse of dimensionality of data classification tasks [4]. In solving the data classification problems, when the feature set is too large, typical approaches will not be able to solve the problem [1]. Different approaches have been proposed to solve the curse of dimensionality in data classification.

In [5] Salimi et.al the feature set sub-selection approach is used to solve the curse of dimensionality. In this research, SVM classifier and with different features subset selection, data classification is performed and the result is compared with artificial neural network method. They have also observed that the SVM classifier has a sensitivity to decrease the number of training samples and feature set and can be used to classify data with the curse of dimensionality issue.

In [6], Yahya is a swarm intelligence-based approach to classify educational data. The effect of the particle swarm optimization approach is investigated to design a classifier using Rocchio Algorithm (RA) to solve the curse of dimensionality issue. By comparing the proposed method to specific data, it can be found that the standard particle swarm optimization algorithm will suffer from poor performance in encountering the stack. Using particle swarm optimization based on the PSO algorithm, the performance of the proposed method is increased in terms of classification accuracy.

In [7], Bach has developed a classifier using a convex neural network to cope with the curse of dimensionality. In the proposed model, geometric interpolation is selected as an active function.

In this research, they show that by defining simple terms for convex functions under the non-convex set of a non-convex set, the same generalization error can be achieved.

Another approach to cope with the curse of dimensionality can be to partition the feature set into multiple features sub-sets so that the classifier problem is solved for each of the feature subsets and finally by using the ensemble classifier, the classification is applied to the entire feature set [8, 9].

In [10], Valentini et al. used bags of SVM and feature selection algorithms to identify malignant melanoma. The results of the study are more accurate or more accurate than the case of SVM classification. It should be noted that in addition to the application of SVM as the classifier, it can be due to the use of feature selection algorithms.

In [11], Jabber et al. have proposed a cluster-based ensemble classification approach to the intrusion detection system. In this research, the k-means clustering method is used and the ensemble classifier is constructed using KNN and ADTree. The simulation results of different data indicate that the accuracy of classification is very high.

Despite the various studies done in this approach, the partition of feature set into features sub-set is still the subject of interest in the literature of this field [12]. In particular [10, 11], refers to high dimensions of feature set for applications such as detecting cancer tumors and intrusion detection systems, and to increasing the exponential difficulty of the curse of the dimensionality of feature set with increasing the number of features. However, in the mentioned studies, a variety of methods have been proposed for partitioning the feature set into the sub-sets, this problem requires more research. In this research, an innovative framework for multi-view ensemble classification is presented. In the proposed method, inspired by the problem of detection of objects in multi-views by humans [13], first, the partitioning of feature set into several views is performed by considering a criterion as a criterion between the features. Clearly, between both features, using a given formula, the value of the two features, which is positive and greater than zero, will be calculated. Then, the collaboration graph is formed based on similarity values between the features. In the collaboration graph, each node is equal to one feature and the relationship between two features is considered as an undirected edge and the value of each feature is considered as the corresponding edge weight. After collaboration graph creation, we use the community detection method to find communities, i.e. views. The community detection method is used to find different communities in which maximize the sum of the weights of the edges of each community and minimize the weight of edges between the two communities. Also, there are no overlapping communities. By finding communities, each community is equal to one view and the features within each community are the features of each view. After this step, for each view by using corresponding training data, a base classifier is trained and a multi-view ensemble classifier is formed from the combination of the base classifiers based on the AdaBoost algorithm. Finally, the classification of the test data is performed with the help of a multi-view ensemble classifier and the accuracy of the classifier is reported. In addition, the proposed method in this research has experimented on real data, i.e. EEG Eye State [14] and two synthetic datasets and the results of the simulation show that classifier accuracy has a reasonable increase compared to typical classifiers.

The proposed method eliminates the disadvantages of previous methods such as removing some features or converting their nature into another space. Furthermore, since the calculation of the collaboration values between the features is independent of one another, it can be done independently and in parallel. In addition, the training of the base classifiers for each view can be done independently and in parallel. Consequently, the proposed method can be performed faster because of the possibility of parallel execution of different parts of it. In addition to increasing the classification accuracy, this method will also reduce the computational cost.

The next chapter reviews the related works in chapter 2. The proposed method is described in detail in chapter 3. In the fourth chapter, the results are presented and the results will be discussed in chapter five.

## 2. Related work

In this chapter, we will study related studies regarding the curse of dimensionality issue. We first categorize the existing solutions and examine the results of each, then propose a proposed solution and evaluate its strengths and weaknesses.

### 2.1. Feature Subset Selection

A part of research related to the issue - resolving the curse of dimensionality problem has focused on reducing the feature set to a limited number of those who have important information. In this approach, it is tried to remove features in data classification without effect or less (redundant features), and the feature set is reduced to a limited number of features that are a subset of the features. Next, we consider different algorithms that have been used to select a subset of features.

In [15], Chandrashekar et.al examined different methods of feature subset selection including filtering, sealing and embedding.

In the filter method [16], variable ratings are used as the main criterion of variable selection by sorting. Ranking methods are used because of their simplicity and good performance for practical applications. A proper scoring criterion is used to score the variables and a threshold to remove the following variables. Since the ranking methods used before the classifier are used to filter related perceptual variables, they are considered the most important ones. An essential feature of a unique feature is that it contains useful information about various objects. This feature can be defined as the feature relevance of feature utility in differentiating between different classes. Here, the relevance of a feature to be discussed is how we can measure the relationship of a feature with objects or output (classes).

The wrapper method [17] uses the predication concept as a black box and the predication performance as a goal function. Since the evaluation of all $2^N$ subsets of variables is an NP-hard problem, they can be found using a heuristic search. As mentioned, we can use several heuristic search algorithms to find a subset of variables. In these algorithms, the goal is to maximize its performance (the classifier performance). Wrapper classification algorithms can be classified into sequential selection algorithms and heuristic search algorithms. The sequential selection algorithms start with an empty set and add the variables to the empty set (from the complete set) to maximize its objective function. These algorithms continue until the objective function reaches its maximum value, continues adding the variables, and finally, when the objective function attains its value, the algorithm stops and announces the set as an optimal subset. To accelerate the selection process of variables, a criterion is chosen that gradually increases the goal and tries to maximize the number of variables to maximize the objective function value. Heuristic search algorithms consider subsets of variables to optimize their target performance. These different subsets are referred to or generated by searching around in the search set or through generating solutions for optimal solutions.

The embedded method [18] is a trade-off between the filter and the wrapper methods that incorporates feature selection in learning the classifier model. Therefore, it can be said that the embedded method, inherits the advantages of the wrapper and filter methods, includes interactions with the learning algorithm, and because they do not require iterative evaluation of feature sets, they are more efficient. The most widely used devices are regulators that aim to find a learning model that is achieved by minimizing the filtering error and keeping the feature coefficients constant (which is a small or zero value). Afterwards, both the adjusted model and the set of features selected are returned as final results.

Other methods of feature subset selection can also be referred to as unsupervised learning [19], genetic algorithm [20] and ensemble feature selection [21].

### 2.2. Feature Set Transformation

In this section, we will examine research related to transforming feature set into another set to reduce its dimensionality and to solve the curse of the dimensionality problem. In this approach, it is attempted to map the feature set into another set using different relationships and transformations. In the new set, the features have a different definition and the number of them can be reduced. In this way, we will solve the data classification with a new feature set in which the curse of dimensionality issue will be resolved by dimensionality reduction. Next, we introduce several approaches to transform the feature set and review the results.

In [22], principal component analysis (PCA) is the most widely used feature set transformation approach. It is a mathematical algorithm that reduces dimensions by maintaining the degree of diversity in the data. This reduction is realized by identifying the directions that are called the main component. The identification of directions can be carried out in such a way that the diversity in the above-mentioned directions must be maximum of its value. Using multiple principal components, each sample can be represented by a relatively small number of features instead of a large number of features. Then we can draw the samples to study their similarities and differences and determine the possibility of classification of samples.

In [23], the method of linear discriminant analysis which is a very common technique for transforming the feature set is introduced. This method is used as a preprocessing step for machine learning and data classification applications. The analysis is usually used as a black box, but it is not sometimes well understood. This technique is developed to transform the properties of the problem into a more complex feature set, which maximizes the ratio of the variance between classes and minimize the variance in classes and thus ensures discrimination. Two types of linear discriminant analysis have to deal with classes: first class-independent and class-dependent. In class-dependent, a multidimensional set is computed for each class to determine its data. While in the class-independent type, each class is considered a separate class against other classes. In this type, there is only an empty set for all classes to determine their data.

In [24], wavelet transform is used as an important analysis tool in general for texture analysis and classification. However, it ignores structural information while providing spectral information in texture images at different scales. In this research, a contextual analysis and classification approach with a linear regression model based on wavelet transform is proposed.

This method works by observing whether there is a distinct correlation between sample images (belonging to the same type of texture) in different frequency regions obtained from the two-dimensional wavelet transform. It is also observed that this relationship varies from tissue to tissue. The linear regression model is used to analyze the correlation and extract textural properties that describe the samples. Therefore, the proposed method not only considers frequency areas but also the correlation between these regions.

In [25], a new feature transformation technique is proposed to improve the accuracy of screening for automatic recognition of hearing pathological sounds. Statistical changes are based on hidden Markov models that are done by obtaining the transform and classification step simultaneously and adjusting model parameters with the criterion that minimizes the classification error. The main feature vector is constructed using short-term and cepstral high-frequency noise parameters. According to the common approaches found in the literature of automatic pathological sounds recognition, the proposed feature set transformation technique shows a significant improvement in performance without adding new features to the original input set. According to the results, it is expected that this technique can provide good results in other areas such as confirming or identifying the speaker.

In [26], a kernel function is proposed to Kernel Principal Components Analysis (KPCA) and Kernel Linear Discriminant Analysis (KLDA). For this purpose, kernel-based functions are combined in linear and non-linear forms by genetic algorithm and genetic programming, respectively. We also use classification error and interaction between features and classes in kernel feature set as evolutionary fitness functions. The proposed method is based on the UCI data set and the Aurora2 database. This study evaluates the methods using clustering and classification accuracy criteria. The experimental results show that KPCA using a non-linear combination of genetic programming-based cores and classification error performance using Gaussian kernel, as well as using a linear combination of cores, has a better performance.

### 2.3. Feature Set Partitioning into Feature Subsets

In this section, we try to study partitioning the feature set into multiple feature subsets to resolve the curse of dimensionality issue and then performing data classification on each feature subset independently. The results of data classification are combined under different sets and results of data classification on the whole feature set. In reviewing related studies in this section, the focus is on the partitioning of feature set into feature subsets and the related works their results will be studied and examined.

In [27], a general framework based on graph processing and mutual information is presented in this study. In this approach, the interaction gain is calculated by normalizing the three-way mutual information criterion as follows:

$$MI(x_i, x_j, y) = \sum_{x_i} \sum_{x_j} \sum_{y} p(x_i, x_j, y) \log \frac{p(x_i, x_j, y) p(x_i) p(x_j) p(y)}{p(x_i, x_j) p(x_i, y) p(x_j, y)} \quad (1)$$

$$IG_{ij} = \frac{MI(x_i, x_j, y)}{H(x_i) + H(x_j)} \quad (2)$$

Eq.1 and Eq.2 are used to compute the three-way mutual information and the interaction gain respectively. In the above equations, $x_i$ denotes the corresponding samples of the $i^{th}$ feature, $x_j$ denotes the corresponding samples of the $j^{th}$ feature, $y$ denotes the corresponding class label of samples of $x_i$ and $x_j$, $p(…)$ representing the probability distribution function, $H(x_i)$ and $H(x_j)$ representing the entropy of $x_i$ and $x_j$ respectively.

In particular, the method is based on clustering similar features to different groups in which some representative features are extracted from them. Two different feature selection techniques are presented based on this framework: first, the simple selection of representative features from the resulting feature groups, and the other is a Meta-heuristic search of music. Comparative experimental evaluation of traditional feature selection techniques on a variety of benchmark datasets shows the efficiency of the proposed approach. With this implementation, significant performance gains can be achieved in terms of overall classification accuracy and dimensionality reduction, while preserving feature semantics and significant redundancy in feature sets.

In [28], a new feature group method of feature selection for intrusion detection is proposed. This method is based on the hierarchical clustering method and is tested on datasets. The hierarchical clustering method is used to construct a hierarchical tree and combines it with mutual information theory. The groups are built with a certain number of hierarchical trees. The largest mutual information between each feature and a class label is then selected in a particular group. Performance evaluation results show that better classification performance can be obtained from such characteristics.

In [29], a feature selection strategy is proposed that uses grouping features to enhance search effectiveness. As a feature selection strategy, this research offers a Variable Neighborhood Search (VNS). Then, the input set is grouped into subsets of features using the concept of Markov blankets. Based on their claim, this is the first time in which Markov's blanket is used to group features. This paper examines the performance of the proposed method by experimenting with 3D datasets from two different domains of microarray and text mining. They are also compared with conventional and competitive techniques. The results show that strategy is a competitive strategy capable of finding a small size of features with the same predictive power as other algorithms used in this study.

In [30], using a combination of hyperspectral data (HS) and LIDAR data for the classification of land cover classification. HS images provide detailed information about the height and LIDAR data for spectral signatures. This paper presents a multiple-fuzzy classifier system for a combination of HS images and LIDAR data. This system is based on fuzzy K - nearest neighbour classification (KNN) from two data sets following the classification of properties on them. Then, the fuzzy decision fusion method is applied to synthesize the results of fuzzy KNN classifiers. The experiments were carried out on the classification of HS images and LIDAR data from Houston, the United States. The proposed fuzzy classifier system for HS images and LIDAR data provides interesting results on the effectiveness and potential of joint use of these data. The combination of the fuzzy classifier in these two data sets improves the classification results compared to single-fuzzy classifiers in each data set.

In [31], a scream-based screening approach for children with autism spectrum disorder (ASD) is introduced that performs early screening and detection. For primary automatic screening, a new classification approach based on defining two types of classifier WSI and SSI is proposed to determine these features and their corresponding samples. In this approach, an exclusive clustering on different datasets based on differentiation of desired features is performed. To test the proposed approach, the dataset of children between 18 to 53 months old, which had been recorded using high-quality sound recording devices and common smartphones in different places such as homes and daily care centers, is used. After pre-processing, the method is used to train WSI and SSI classifiers using data related to 10 boys with ASD and 10 normal boys. The pre-trained classifier is trained on 14 boys and 7 girls with ASD and 14 normal boys and 7 normal girls. The sensitivity, specificity, and accuracy of the proposed approach for boys were 85.71%, 100%, and 92.85 respectively. These measurements for girls are 71.42%, 100% and 85.71%, respectively. Based on the obtained results, the proposed approach outperforms current classification methods.

## 2.4. Conclusions

In this chapter, we have investigated the proposed methods to solve the curse of dimensionality of feature set in data classification problems. As mentioned previously, the focus is on three different approaches in the literature:

1) Feature Subset Selection: in this approach, the feature set is reduced to several limited features that have important information. In this way, the curse of dimensionality is resolved and the data classification problem can be achieved on the reduced feature set.

2) Feature Set Transformation: in this approach, the feature set is transformed into another set so that the dimensionality of the feature set will decrease and the curse of dimensionality of the feature set is resolved. One of the most important approaches in this field is the Principal Component Analysis (PCA) which is the oldest and most widely used method for transforming features set. Due to the transformation of the feature set, its dimensions decrease and in most cases the data classification results using the transformed set is desirable. The disadvantage of this method is to change the real interpretation of the feature set to other features that do not necessarily mean any real meaning.

3) Feature Set Partitioning into Feature Subsets: in this approach, the feature set is partitioned into different feature subsets. It is possible to partition the feature set into subsets in a way that the features in each subset have a maximum relationship to each other and they will have a minimal relationship with other features of different subsets. Each subset is then classified in parallel and the results of each subset are combined with a data fusion method. The results obtained from the combination of base classification results is equivalent to solving the data classification on the whole feature set. The main advantage of this method is to preserve the nature of the features and prevent them from eliminating which is the previous methods' disadvantage. Some studies [27, 28, 29] have shown that the classification results with this approach can even lead to better results than the two previously mentioned methods.

## 3. The Proposed Method

As mentioned in the introduction, in this section, we consider the proposed method to resolve the curse of dimensionality issue in the data classification problem. The proposed approach is inspired by multiple views theory [13]. According to this theory, when humans classify objects, they examine objects' different views and therefore, by combining the results of different views examination, they will result in object recognition. Inspired by this theory, the proposed method is based on partitioning feature set into several feature subsets so that each subset is considered as a view. Then, a base classifier is trained for each view (sub-set) by using its corresponding training data. Finally, the base classifiers are combined based on the AdaBoost algorithm and form a multi-view ensemble classifier. Finally, the multi-view ensemble classifier classifies the test data and the accuracy is reported. The framework of the proposed method, which is a framework for multi-view ensemble classification, is shown in Fig.1.

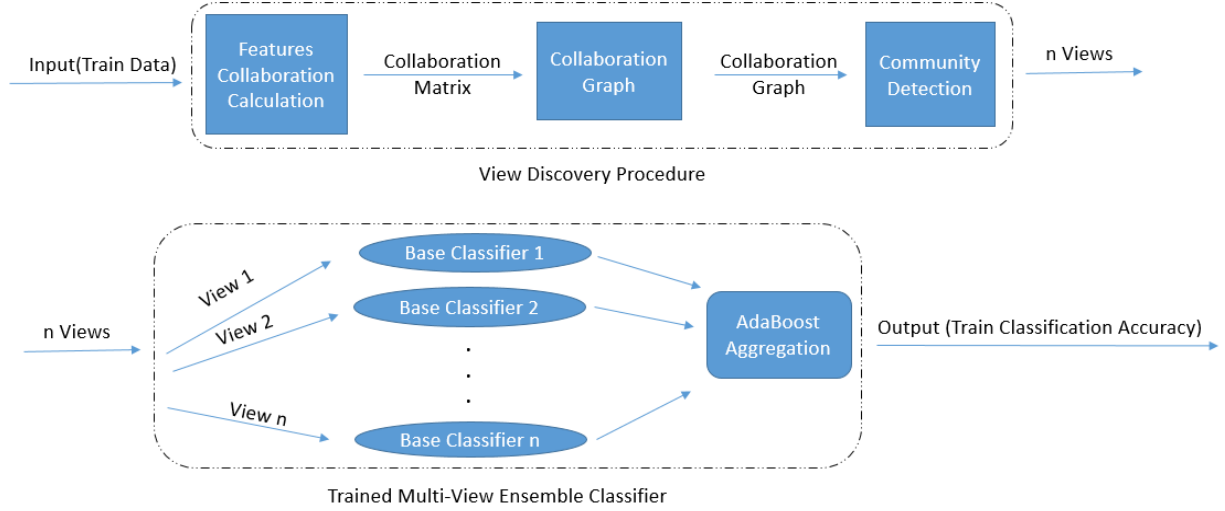

Figure 1: The framework for multi-view ensemble classification

We first define the collaboration value of each two features and then define the proposed method based on the defined criterion.

Suppose that $f(x;\theta)$ is a classifier of the parameter $\theta$ and the samples input of the feature $x$, we define the error $e$ as follows:

$$e = \min_{\theta} E[(f(x;\theta) - y)^2] \qquad (3)$$

In Eq.3, $E$ is the expectation value and $y$ is the class label of corresponding samples of features $x$. To be more specific, the error value is equal to the amount of mathematical expectation of the squared distance between the predicted class label and the actual class label over the parameter $\theta$.

Now using Eq.3, we define the errors $e_1$, $e_2$, and $e_{1,2}$, as follows:

$$e_1 = \min_{\theta} E[(f(x_1;\theta) - y)^2] \qquad (4)$$

$$e_2 = \min_{\theta} E[(f(x_2;\theta) - y)^2] \qquad (5)$$

$$e_{1,2} = \min_{\theta} E[(f(x_1, x_2;\theta) - y)^2] \qquad (6)$$

$$\text{If } e_{1,2} > \min(e_1, e_2) \rightarrow e_{1,2} = \min(e_1, e_2) \tag{7}$$

In Eq.6, $f(x_1, x_2; \theta)$ the classifier with the parameter $\theta$ and the samples input of the feature $x_1$ and $x_2$. Also, $y$ is the class label of the corresponding samples of features $x_1$ and $x_2$. Also, as indicated in Eq.7, if the amount of error obtained by applying both the features $x_1$ and $x_2$ simultaneously is more than the minimum error obtained by applying the feature $x_1$ and the feature $x_2$ independently, $e_{1,2}$ is equal to the minimum error obtained by applying the feature $x_1$ and the feature $x_2$ independently. Based on the above-defined equations, the collaboration value of each two features is expressed as follows:

$$colab(x_1, x_2) = \min(e_1, e_2) - e_{1,2} \tag{8}$$

$$0 \leq e_{1,2} \leq \min(e_1, e_2) \tag{9}$$

$$0 \leq colab(x_1, x_2) \leq \min(e_1, e_2) \tag{10}$$

In Eq.8, $colab(x_1, x_2)$ shows the collaboration value of two features $x_1$ and $x_2$. It should be noted that the higher the amount of collaboration is, the two features are working together to classify the data better. The upper and lower bounds of $e_{1,2}$ and $colab(x_1, x_2)$ are also shown in Eq.9 and Eq.10, respectively.

Based on the collaboration value of each two features, the features collaboration graph that is an undirected weighted graph is defined as a tuple $G(V, E)$. In the graph, the $V$ represents the set of graph nodes so that each node represents a feature. The $E$ will represent the set of graph edges so that if the collaboration value of two features is non - zero, there would be an edge between the two corresponding features such that the edge weight is equal to the collaboration value of two corresponding features.

According to the definition of collaboration value of two features and how to form the collaboration graph, the proposed method steps are expressed as follows:

1) For every two features, we calculate the collaboration value of the two features using Eq.8.
2) According to the collaboration values of each two features, we form the features collaboration graph, namely $G\ (V, E)$.
3) By implementing the community detection method on the collaboration graph of features, graph communities are found in such a way that the edges between every two nodes within each community have the highest weight and the edges connected between the nodes of the two different communities have the lowest possible weight. It should be noted that in this research, the community detection method works such that communities do not have any overlaps. It should also be noted that the communities obtained at this step will be the views of the problem.

4) For each view, we train a base classifier using the corresponding training data of the view.
5) The base classifiers are combined based on the AdaBoost algorithm and the multi-view ensemble classifier is formed.
6) The classification of the test data using the multi-view ensemble classifier is performed and the classification accuracy is reported.

To further explain the proposed method, a simple problem is presented in the form of an illustration. We assume that the dataset has six features. After calculating the collaboration value of each two features, the features collaboration matrix can be obtained as follows:

$$\begin{array}{c} \phantom{1}\begin{array}{cccccc} 1 & 2 & 3 & 4 & 5 & 6 \end{array} \\ \begin{array}{c} 1 \\ 2 \\ 3 \\ 4 \\ 5 \\ 6 \end{array} \begin{bmatrix} 0.00 & 0.10 & 0.07 & 0.11 & 0.00 & 0.00 \\ 0.10 & 0.00 & 0.12 & 0.10 & 0.08 & 0.06 \\ 0.07 & 0.12 & 0.00 & 0.10 & 0.00 & 0.00 \\ 0.11 & 0.10 & 0.10 & 0.00 & 0.00 & 0.00 \\ 0.00 & 0.08 & 0.00 & 0.00 & 0.00 & 0.06 \\ 0.00 & 0.06 & 0.00 & 0.00 & 0.06 & 0.00 \end{bmatrix} \end{array}$$

Figure 2: The features collaboration matrix

As observed in Fig.2, the collaboration value of each two features is represented as a real number greater than or equal to zero. For example, the collaboration value of the second and third features is equal to 0.12. It should be noted that since the arrangement of the features does not affect the calculation of the collaboration value of each two features, the features collaboration matrix will be symmetric. Also, the collaboration value of a feature with itself is not useful, so the main diagonal of the matrix is zero.

After calculating the features collaboration matrix, the features collaboration graph is drawn as follows:

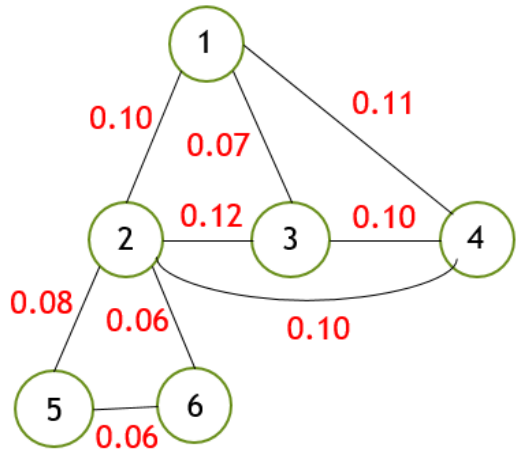

Figure 3: The features collaboration graph

As shown in Fig.3, the features collaboration graph, which is an undirected weighted graph, includes six nodes representing six features. Also, between every two features, whose collaboration value is non-zero, there would be an edge such that the edge weight is equal to the collaboration value of the two features. After drawing the features collaboration graph, using the community detection method, the graph communities or the views, can be obtained as follows:

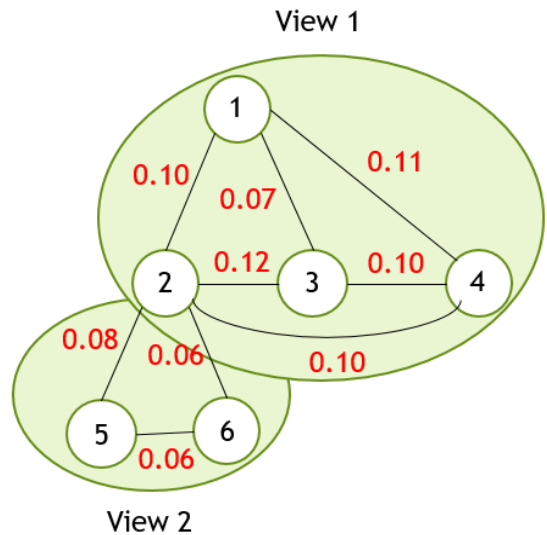

Figure 4: The views found in the features collaboration graph using the community detection method

As shown in Fig.4, two views (communities) for the collaboration graph are found. Therefore, the problem views will be as follows:

View 1 = {1, 2, 3, 4}

View 2 = {5, 6}

The proposed method algorithms are as follows:

**Algorithm1 ViewDiscoveryProcedure** (Train Dataset):

1. $n \leftarrow$ number of samples
2. $f \leftarrow$ number of features
3. $V \leftarrow$ features set
4. $E \leftarrow \{\}$
5. **for** $i \leftarrow 1\ to\ f$
6.     **for** $j \leftarrow i + 1\ to\ f$
7.         $colab\_matrix[i,j] \leftarrow$ **calculate** $colab(x_i, x_j)$
8.         **If** $colab\_matrix[i,j]$ is not zero
9.             $E \leftarrow E \cup \{(i,j, colab\_matrix[i,j])\}$
10.     **endfor**
11. **endfor**
12. $colab\_graph \leftarrow$ **create** $G(V, E)$
13. $views \leftarrow$ **communitydetection** $(colab\_graph)$
14. **return** $views$

**Algorithm2 Multi-ViewEnsembleClassifierTraining** (Train Dataset, Test Dataset, Views)

1. $base\_classifiers \leftarrow \{\}$
2. **for** each $view$ in $views$
3.     $trained\_base\_classifier \leftarrow$
4.         **train $baseclassifier$** ($view$ and its corresponding train samples)
5.     $base\_classifiers \leftarrow base\_classifiers \cup trained\_base\_classifier$
6. **endfor**
7. $multiview\_ensemble\_classifier \leftarrow$
8.    **fuse** $base\_classifiers$ based on the **Adaboost aggregation**
9. $accuracy \leftarrow$
      **test** $mutiview\_ensemble\_classifier$ ($views$ and their corresponding test samples)
10. **return** $accuracy$

In Algorithm 1, the steps for finding the problem views are described. In detail, in lines 1-12, the collaboration value of each two features is calculated and the features collaboration graph is formed based on it. In line 13, using the community detection method, the problem views are found and returned as the output of Algorithm 1.

In Algorithm 2, having the problem views and the training and test dataset, the multi-view ensemble classifier is formed and the accuracy of classification on the test data is reported. In detail, in lines 1-6, base classifiers are trained for each view using their corresponding training data. In line 7, the base classifiers are combined based on the AdaBoost algorithm and form the multi-view ensemble classifier. In line 9, the classification of the test data is performed using the multi-view ensemble classifier and the accuracy obtained in line 10 is returned as the output of Algorithm 2.

It should be noted that according to Algorithm 1, the time complexity of the collaboration value calculation of each two features and the formation of the features collaboration graph would be $O(f^2)$ where $f$ is the number of features of the data classification task. It is obvious that the collaboration value calculation of features is polynomial and can be done independently. Therefore, using parallel programming, the collaboration value calculations can be done simultaneously with the minimum time required.

In the next chapter, we discuss and analyze the simulation results of the proposed method on real and synthetic datasets.

## 4. Experimental Results

In this chapter, we will examine various experiments and analyze the results. Before evaluating the design experiments and the results obtained through them, we describe and analyze the datasets used in this section.

To experiment with the proposed method, we use a real dataset and two synthetic datasets.

The real dataset is named EEG Eye State obtained from performing EEG on different individuals in the case that their eyes are opened or closed [14]. The EEG duration has been 117 seconds and the eye state of individuals has been recorded by a camera during this period. After performing various processing, 15 features are extracted, each of which contains parts of the EEG, as well as the eye state of individuals, i.e., number one for an opened eye and number zero for a closed eye. The experiment has been done on 14980 individuals.

The synthetic dataset is constructed using the Scikit-learn library in a python environment [32]. The library receives several input parameters including the number of samples, the number of features, and the number of informative features and then generates the datasets based on them.

In Tab.1, the specifications of the real and synthetic datasets are given:

Table 1: The specification of the real and synthetic datasets

| Dataset | Number of Samples | Number of Features | Number of Classes |
|---|---|---|---|
| **EEG Eye State** | 14980 | 15 | 2 |
| **Synthetic 1** | 1000 | 20 | 2 |
| **Synthetic 2** | 10000 | 10 | 2 |

To simulate the proposed method on the real and synthetic datasets, the following assumptions have been applied:

1) All of the classifiers for the collaboration value calculation of each two features and also as the base classifiers are the SVM classifier.
2) We use the ASLPAw algorithm [33] as the community detection method.
3) In the simulation of different parts of the proposed method, the same Random Seed value is used to counteract the random effect of the results due to the randomness of the methods.
4) The simulations have been done using a computer with Core i5 2.4 GHz CPU and 8GB DDR3 RAM and on Windows.

It should also be noted that the simulation results are compared with the results obtained by the method presented in [27] which uses the interaction gain (normalized three-way mutual information).

In Tab.2 and Fig.5, the simulation results of the proposed method are shown on different datasets:

Table 2: The classification accuracy achieved by different criteria and the exhaustive search

| Dataset | Classification Accuracy Using Whole Feature Set | Classification Accuracy Using Interaction Gain | Classification Accuracy Using Collaboration Value of Two Features | Classification Accuracy Using Exhaustive Search |
|---|---|---|---|---|
| EEG Eye State | 55.57% | 59.01% | 62.18% | 65.23% |
| Synthetic 1 | 98.50% | 70.00% | 99.00% | 99.00% |
| Synthetic 2 | 95.50% | 90.30% | 95.50% | 97.80% |

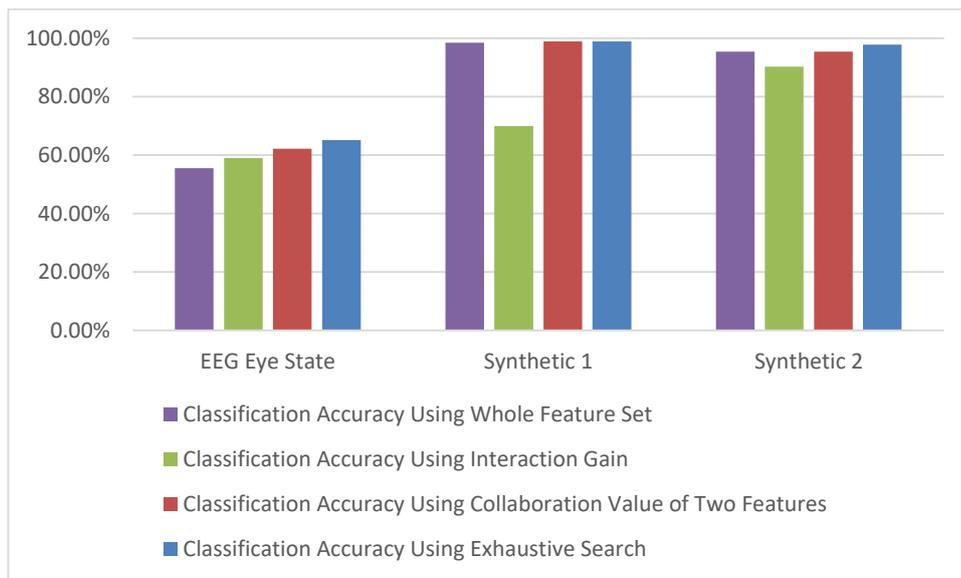

Figure 5: The classification accuracy achieved by different criteria and the exhaustive search

In the reported outcomes of Tab.2, the first column indicates the classification accuracy by applying a single classifier and using the whole feature set altogether. In the second column, the classification accuracy of this study was reported using the interaction gain criterion and in the third column was reported using the collaboration value of each two features. In the last column, the classification accuracy of the proposed method is obtained using the exhaustive search on all possible problem views and the maximum accuracy of the obtained results is reported.

The results in Tab.2 and Fig. 5 show that the proposed method achieves the maximum classification accuracy in all three datasets. In particular, in the synthetic 1 dataset, it can be observed that when most of the features collaborate and the classification accuracy of using all features is 98.5%, the proposed method not only detects the features that are collaborating and place them into the same view but also detects few non-collaborating features that should be in different views. On the other hand, in the synthetic 1 dataset, the interaction gain [27], as the partitioning criterion of features set into features subsets, has not been able to distinguish the collaboration of different features correctly and caused them to be placed in different subsets (views) and consequently reduces the classification accuracy to 70%. Therefore, it can be concluded that in all of the mentioned datasets, the proposed method using the collaboration value of two features could be able to partition the feature set into different views and achieve a better accuracy using the multi-view ensemble classifier.

Moreover, comparing the results obtained from the classification accuracy of the proposed method using the collaboration value of two features and the exhaustive search, it can be concluded that the views found by the proposed method are fairly similar to the views found by the exhaustive search and hence the classification accuracy of both is closely similar. Therefore, the proposed method works very well in finding near-optimal views.

## 5. Conclusion

In this chapter, we discuss the conclusion of the proposed method and the possible future works. As described in the previous chapters, in this research an innovative framework for multi-view ensemble classification is proposed. This method is based on the introduction of a new criterion based on the collaboration value of each two features. Then, based on the calculated collaboration of each two features, the features collaboration graph, which is an undirected weighted graph, is formed. For the formation of an undirected weighted graph, each feature will be considered as a node and there would be an edge between every two nodes if the collaboration value of the two corresponding features is non-zero. The weight of each edge is equal to the collaboration value of the two corresponding features. After the formation of the collaboration graph of features, using the community detection method, graph communities (without any overlaps) are detected in such a way that each community is considered as a view (subset). Then, a base classifier is trained for each view using its corresponding training data. The base classifiers are then combined based on the AdaBoost algorithm and form the multi-view ensemble classifier. Finally, using the test dataset, the multi-view ensemble classifier is tested and the accuracy is reported.

According to the results of chapter 4 of this research, it can be observed that the proposed method is more accurate than the related methods on real and synthetic datasets. It should be noted that even in the datasets where most of the features are well-collaborated, the proposed method identifies the features of each view properly using the collaboration value of each two features. Obviously, in determining the views, the proposed method can perform better and more accurately. One of the disadvantages of the proposed method is the much time needed for collaboration value calculation of each two features in the training step. Since the training step of the classifier is performed only once and also it is possible to do the collaboration calculation parallel and fast so that the time-consuming issue of the proposed method can be resolved.

For future works, we can consider the following to work on more:

1) For the calculation of the collaboration value of each of the two features, the type of classifier can be selected based on the intrinsic characteristics of the two features to achieve better collaboration values such that the classifier is suitable for two features.
2) We can determine different base classifiers for different views according to the intrinsic characteristics of each view's features to increase the classification accuracy, as well as, reducing the computational cost of the base classifier.
3) Although the formation of the features collaboration graph is performed only once, it is possible to investigate the solutions that accelerate this step.
4) Views can be determined to overlap each other and this overlap in some cases will probably increase the classification accuracy.
5) We can use different algorithms as community detection methods to find the views that lead to higher classification accuracy.
6) The combination algorithms of base classifiers can be studied and if there would be any algorithms, rather than the AdaBoost algorithms that can lead to higher classification accuracy, should be considered.

7) The analysis of the time complexity of the proposed method can be studied in more detail.
8) The theoretical analysis of the proposed method to obtain error bounds for the method can be investigated.
9) The proposed method must be tested on more real and synthetic datasets using different algorithms and classifiers.
10) The proposed method can be designed and optimized for a particular application such as an Intrusion Detection System (IDS) to achieve better performance than existing methods.

**References**

[1] Baggenstoss, Paul M. "Class-specific classifier: avoiding the curse of dimensionality." *IEEE Aeroset and Electronic Systems Magazine* 19, no. 1 (2004): 37-52.

[2] Duda, Richard O., and Peter E. Hart. *Pattern classification and scene analysis*. Vol. 3. New York: Wiley, 1973.

[3] Zhang, Guoqiang Peter. "Neural networks for classification: a survey." *IEEE Transactions on Systems, Man, and Cybernetics, Part C (Applications and Reviews)* 30, no. 4 (2000): 451-462.

[4] Friedman, Jerome H. "On bias, variance, 0/1—loss, and the curse-of-dimensionality." *Data mining and knowledge discovery* 1, no. 1 (1997): 55-77.

[5] Salimi, Amir, Mansour Ziaii, Ali Amiri, Mahdieh Hosseinjani Zadeh, Sadegh Karimpouli, and Mostafa Moradkhani. "Using a feature subset selection method and support vector machine to address the curse of dimensionality and redundancy in Hyperion hyperspectral data classification." *The Egyptian Journal of Remote Sensing and Set Science* 21, no. 1 (2018): 27-36.

[6] Yahya, Anwar Ali. "Swarm intelligence-based approach for educational data classification." *Journal of King Saud University-Computer and Information Sciences* 31, no. 1 (2019): 35-51.

[7] Bach, Francis. "Breaking the curse of dimensionality with convex neural networks." *The Journal of Machine Learning Research* 18, no. 1 (2017): 629-681.

[8] Chan, Aki PF, Patrick PK Chan, Wing WY Ng, Eric CC Tsang, and Daniel S. Yeung. "A novel feature grouping method for ensemble neural network using localized generalization

error model." *International Journal of Pattern Recognition and Artificial Intelligence* 22, no. 01 (2008): 137-151.

[9] Xu, Tianpei, Ying Ma, and Kangchul Kim. "Telecom Churn Prediction System Based on Ensemble Learning Using Feature Grouping." *Applied Sciences* 11, no. 11 (2021): 4742.

[10] Valentini, Giorgio, Marco Muselli, and Francesca Ruffino. "Cancer recognition with bagged ensembles of support vector machines." *Neurocomputing* 56 (2004): 461-466.

[11] Jabbar, M. Akhil, Rajanikanth Aluvalu, and S. Sai Satyanarayana Reddy. "Cluster-based ensemble classification for the intrusion detection system." In *Proceedings of the 9th International Conference on Machine Learning and Computing*, pp. 253-257. 2017.

[12] Gan, Guojun, and Michael Kwok-Po Ng. "Subset clustering with automatic feature grouping." *Pattern Recognition* 48, no. 11 (2015): 3703-3713.

[13] Tarr, Michael J., and Heinrich H. Biilthoff. "Is Human Object Recognition Better Described by Geon Structural Descriptions or by Multiple Views? Comment on Biederman and." (1995).

[14] A. Frank and A. Asuncion, "UCI machine learning repository," 2010, http://archive.ics.uci.edu/ml.

[15] Chandrashekar, Girish, and Ferat Sahin. "A survey on feature selection methods." *Computers & Electrical Engineering* 40, no. 1 (2014): 16-28.

[16] Sánchez-Maroño, Noelia, Amparo Alonso-Betanzos, and María Tombilla-Sanromán. "Filter methods for feature selection–a comparative study." In *International Conference on Intelligent Data Engineering and Automated Learning*, pp. 178-187. Springer, Berlin, Heidelberg, 2007.

[17] Chen, Gang, and Jin Chen. "A novel wrapper method for feature selection and its applications." *Neurocomputing* 159 (2015): 219-226.

[18] Wang, Suhang, Jiliang Tang, and Huan Liu. "Embedded unsupervised feature selection." In *Proceedings of the AAAI Conference on Artificial Intelligence*, vol. 29, no. 1. 2015.

[19] Mitra, Pabitra, C. A. Murthy, and Sankar K. Pal. "Unsupervised feature selection using feature similarity." *IEEE transactions on pattern analysis and machine intelligence* 24, no. 3 (2002): 301-312.

[20]     Amini, Fatemeh, and Guiping Hu. "A two-layer feature selection method using genetic algorithm and elastic net." *Expert Systems with Applications* 166 (2021): 114072.

[21]     Saeys, Yvan, Thomas Abeel, and Yves Van de Peer. "Robust feature selection using ensemble feature selection techniques." In *Joint European Conference on Machine Learning and Knowledge Discovery in Databases*, pp. 313-325. Springer, Berlin, Heidelberg, 2008.

[22]     Ringnér, Markus. "What is principal component analysis?." *Nature Biotechnology* 26, no. 3 (2008): 303-304.

[23]     Tharwat, Alaa, Tarek Gaber, Abdelhameed Ibrahim, and Aboul Ella Hassanien. "Linear discriminant analysis: A detailed tutorial." *AI communications* 30, no. 2 (2017): 169-190.

[24]     Wang, Zhi-Zhong, and Jun-Hai Yong. "Texture analysis and classification with linear regression model based on wavelet transform." *IEEE transactions on image processing* 17, no. 8 (2008): 1421-1430.

[25]     Arias-Londoño, Julián D., Juan I. Godino-Llorente, Nicolás Sáenz-Lechón, Víctor Osma-Ruiz, and Germán Castellanos-Domínguez. "An improved method for voice pathology detection by means of a HMM-based feature set transformation." *Pattern recognition* 43, no. 9 (2010): 3100-3112.

[26]     Zamani, Behzad, Ahmad Akbari, and Babak Nasersharif. "Evolutionary combination of kernels for nonlinear feature transformation." *Information Sciences* 274 (2014): 95-107.

[27]     Zheng, Ling, Fei Chao, Neil Mac Parthaláin, Defu Zhang, and Qiang Shen. "Feature grouping and selection: A graph-based approach." *Information Sciences* 546 (2021): 1256-1272.

[28]     Song, Jingping, Zhiliang Zhu, and Chris Price. "Feature grouping for intrusion detection system based on hierarchical clustering." In *International Conference on Availability, Reliability, and Security*, pp. 270-280. Springer, Cham, 2014.

[29]     García-Torres, Miguel, Francisco Gómez-Vela, Belén Melián-Batista, and J. Marcos Moreno-Vega. "High-dimensional feature selection via feature grouping: A Variable Neighborhood Search approach." *Information Sciences* 326 (2016): 102-118.

[30]     Bigdeli, Behnaz, Farhad Samadzadegan, and Peter Reinartz. "Feature grouping-based multiple-fuzzy classifier system for fusion of hyperspectral and LIDAR data." *Journal of Applied Remote Sensing* 8, no. 1 (2014): 083509.

[31]     Khozaei, Aida, Hadi Moradi, Reshad Hosseini, Hamidreza Pouretemad, and Bahareh Eskandari. "Early screening of autism spectrum disorder using cry features." PloS one 15, no. 12 (2020): e0241690.

[32]     [Scikit-learn: Machine Learning in Python](), Pedregosa *et al.*, JMLR 12, pp. 2825-2830, 2011.

[33]     Rossetti, Giulio, Letizia Milli, and Rémy Cazabet. "CDLIB: a python library to extract, compare and evaluate communities from complex networks." *Applied Network Science* 4, no. 1 (2019): 1-26.